\documentclass{article}
\usepackage{ijcai20}

\usepackage{times}
\usepackage{soul}
\usepackage{url}
\usepackage[hidelinks]{hyperref}
\usepackage[utf8]{inputenc}
\usepackage[small]{caption}
\usepackage{graphicx}
\usepackage{amsmath}
\usepackage{amsthm}
\usepackage{booktabs}
\usepackage{algorithm}
\usepackage{algorithmic}
\usepackage{multirow}
\usepackage{subfig}
\usepackage{subfloat}
\usepackage{amsfonts}
\def\Mat#1{{\boldsymbol{#1}}}
\def\Vec#1{{\boldsymbol{#1}}}

\newtheorem*{definition}{Definition}

\makeatletter
\DeclareRobustCommand\onedot{\futurelet\@let@token\@onedot}
\def\@onedot{\ifx\@let@token.\else.\null\fi\xspace}

\makeatother

\title{Self-Supervised Graph Representation Learning via Global Context Prediction}

\author{
Zhen Peng$^1$\and
Yixiang Dong$^1$\and
Minnan Luo$^1$\and
Xiao-Ming Wu$^2$\and
Qinghua Zheng$^1$
\affiliations
$^1$MOEKLINNS, School of Computer Science and Technology,
Xi'an Jiaotong University, China\\
$^2$Department of Computing, The Hong Kong Polytechnic University, Hong Kong
\emails
zhenpeng27@outlook.com, 
dyx1102@stu.xjtu.edu.cn, 
minnluo@xjtu.edu.cn, 
xiao-ming.wu@polyu.edu.hk,
qhzheng@xjtu.edu.cn
}

\begin{document}

\maketitle

\begin{abstract}
To take full advantage of fast-growing unlabeled networked data, this paper introduces a novel self-supervised strategy for graph representation learning by exploiting natural supervision provided by the data itself. Inspired by human social behavior, we assume that the global context of each node is composed of all nodes in the graph since two arbitrary entities in a connected network could interact with each other via paths of varying length. Based on this, we investigate whether the global context can be a source of free and effective supervisory signals for learning useful node representations. Specifically, we randomly select pairs of nodes in a graph and train a well-designed neural net to predict the contextual position of one node relative to the other. Our underlying hypothesis is that the representations learned from such within-graph context would capture the global topology of the graph and finely characterize the similarity and differentiation between nodes, which is conducive to various downstream learning tasks. Extensive benchmark experiments including node classification, clustering, and link prediction demonstrate that our approach outperforms many state-of-the-art unsupervised methods and sometimes even exceeds the performance of supervised counterparts.
\end{abstract}

\section{Introduction}

Graph representation learning has attracted a great deal of interest from researchers in recent years. Learning effective node representations can benefit a variety of practical downstream tasks, \emph{e.g.}, classification~\cite{chen2018fastgcn}, community detection~\cite{ye2018deep}, and graph alignment~\cite{heimann2018regal}. Compared to many well-performed supervised algorithms~\cite{huang2018adaptive,xu2018graph}, unsupervised methods  ~\cite{grover2018graphite,velickovic2018deep} have a definite advantage that they are free from expensive manual labeling effort and therefore can fully utilize a vast amount of unlabelled data. However, despite the empirical success, \textit{what} should be learned has been a central issue for unsupervised learning. In the absence of handcrafted annotation, how to design an appropriate objective function to learn desirable node representations is a challenging problem.

\begin{figure}[t]
		\setlength{\belowcaptionskip}{-0.2cm}
		\centering
		{\includegraphics[width=0.45\textwidth]{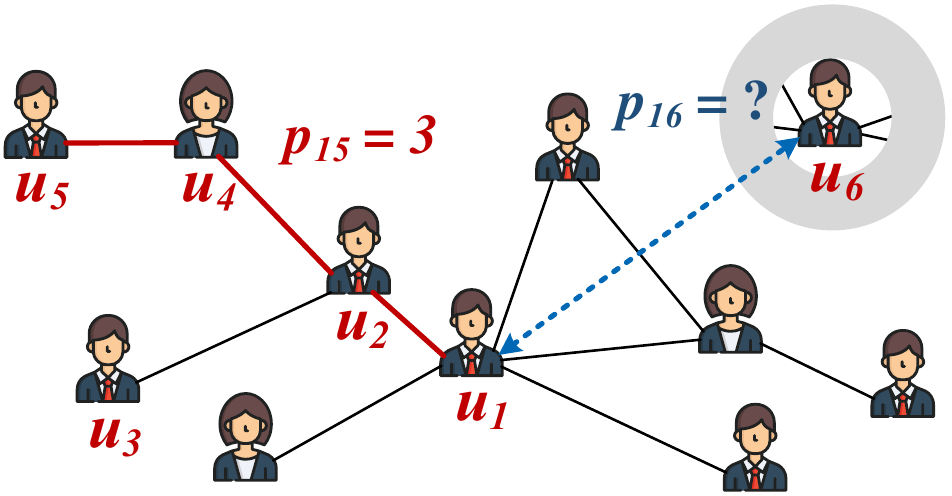} 
		\caption{A toy example of our self-supervised task involving predicting the relative contextual position of one node to another.}
		\label{fig:g1}}
\end{figure}

Fortunately, self-supervised learning~\cite{jing2019self}, as a branch of unsupervised learning, empowers us to train unlabeled data with free supervised signals obtained from the data itself. It has been successfully applied on image and video data~\cite{doersch2015unsupervised,gidaris2018unsupervised}. By training various pretext tasks such as predicting the rotation angle of an image or inferring the correct temporal order of a sequence of frames in a video, useful latent vectors can be learned from unlabeled data in a supervised manner, which helps to achieve desirable performance in relevant tasks such as object detection and classification.  

\begin{figure}[t]
		\setlength{\belowcaptionskip}{-0.2cm}
		\centering
		{\includegraphics[width=0.45\textwidth]{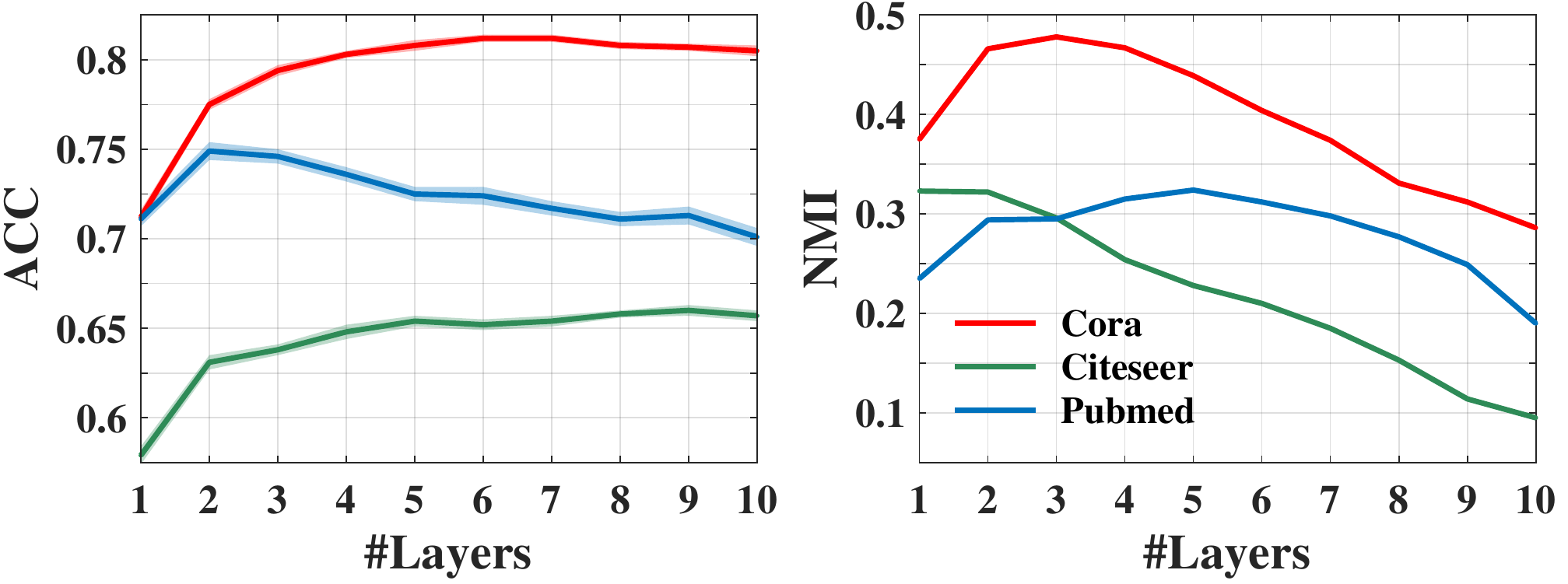}
		\caption{Results of an unsupervised baseline formulated by stacking varying number of graph convolutional layers on node classification (\textbf{left}) and clustering (\textbf{right}) tasks. Initially the performance improves as the number of layer increases. But more layers lead to over-smoothing and performance decay.}
		
		\label{fig:ga}}
\end{figure}

A natural question is: can we also get free supervision from graph-structured data? After much deliberation, our answer is positive. Recall a common phenomenon in a social network as illustrated in Figure~\ref{fig:g1}, you ($u_{1}$) are more likely to interact with your direct friends (\emph{e.g.}, $u_{2}$) than your friends' friends (\emph{e.g.}, $u_{4}$), but it is also possible that your friends (\emph{e.g.}, $u_{2}$) may be influenced by their other friends (\emph{e.g.}, $u_{3}$ and $u_{4}$) and then affect you, due to the link structure of the network. In this sense, for a node $v_{i}$ in a graph $\mathcal{G}$, all nodes in $\mathcal{G}$ constitute its context as any other node may interact with it through a path of varying length\footnote{Typically, the length of a path refers to the number of edges it traverses, also known as \textbf{hop count}. If there is no path between two nodes, their distance is infinite.}, \emph{e.g.}, $u_{1}$ interacts with $u_{5}$ via a path of length 3. Since such context covers the global topology of the graph, we call it the global context. Nevertheless, how to effectively capture the global structure of a graph remains a challenging issue. Although existing graph neural network methods such as graph convolutional networks (GCNs)~\cite{kipf2016semi} can stack multiple layers to capture high-order relations between nodes, they suffer from over-smoothing when the number of layers increases~\cite{li2018deeper}, as illustrated in Figure~\ref{fig:ga}. Furthermore, it is difficult to choose an appropriate number of stacked layers. In this work, we propose to use the length of a path, \emph{i.e.}, hop count, to characterize the global context. The path length can naturally and faithfully reflect the extent of similarity between two nodes. The shorter the path length, the greater the degree of interaction between them. More importantly, such supervisory information can be obtained for free from the graph data, making it possible to learn node representations in a self-supervised fashion.




This paper provides a \underline{s}elf-\underline{s}upervised \underline{g}raph \underline{r}epresentation \underline{l}earning framework S$^{2}$GRL involving predicting relative contextual position for a pair of nodes in a graph. In particular, given two arbitrary nodes, the task is to train a neural net to infer the contextual position of one node relative to the other. For instance, in Figure~\ref{fig:g1}, the neural net should be able to answer a question: is $u_{6}$ one hop away from $u_{1}$, or two or more hops away? To perform well on this task, it requires the learned node representations to encode global topological information while capable of discriminating the similarity and dissimilarity between pairs of nodes. The main contributions of our work are summarized as follows:

\begin{itemize}
		\item We make the first attempt to investigate a natural supervisory signal hidden in graph-structured data, \emph{i.e.}, hop count, and exploit this signal to learn node representations on unlabeled datasets in a self-supervised manner.
		\item We propose an effective self-supervised learning framework S$^{2}$GRL that trains a neural net to predict the relative contextual position between pairs of nodes, which learns global-context-aware node representations.
		\item We conduct extensive experiments to evaluate S$^{2}$GRL on three common learning tasks. The results show that it exhibits competitive performance compared with state-of-the-art unsupervised methods and sometimes even outperforms some strong supervised baselines.
\end{itemize}

\section{Related Work}
\textbf{Self-supervised learning} is a form of unsupervised learning where the data provides the supervision to train a pretext task. The key is to automatically generate supervisory signals based on the data itself, which helps to guide the learning algorithm to capture the underlying patterns of the data. As a general technique, self-supervised learning finds many applications, ranging from language modeling~\cite{wu2019self} to robotics~\cite{jang2018grasp2vec}. Notably, it has been widely used on image and video data to learn useful visual features. Various pretext tasks have been proposed such as cross-channel prediction, spatial context prediction, colorization, and watching objects move~\cite{jing2019self}. Although self-supervision has been successfully applied in many areas, it remains unclear whether it works or not in the graph domain. The goal of this paper is to investigate its effectiveness in learning on graph-structured data.

\noindent \textbf{Graph representation learning} is an important task and has become a research hot-spot in recent years. In general, existing approaches are broadly categorized as (1) factorization-based~\cite{qiu2018network}, (2) random walk-based~\cite{perozzi2014deepwalk}, and (3) neural network-based~\cite{li2018adaptive}. 

Recently, graph convolutional network (GCN)~\cite{kipf2016semi} and its multiple variants have become the dominant approach in graph modeling, thanks to the utilization of graph convolution that effectively fuses graph topology and node features. However, the majority of these methods~\cite{velivckovic2017graph,zhang2018gaan,xu2018graph} requires external guidance, \emph{i.e.}, annotated labels, which limits their applicability. In contrast, unsupervised algorithms~\cite{hamilton2017inductive,grover2018graphite,velickovic2018deep} do not require any external labels, but their performances are often not comparable to the supervised counterparts. Some unsupervised methods require advanced knowledge and sophisticated design to ensure their models can learn useful node representations without explicit supervision. Fortunately, self-supervised learning opens up an opportunity for betting utilizing the abundant unlabeled data. A recently proposed multi-stage self-supervised framework M3S~\cite{sun2019multi} has been shown empirically successful. However, in essence, M3S does not get rid of external guidance as it still requires a few initial labels as the basis for enlarging the label set subsequently. To make better use of unlabeled data, in this work, we propose a novel self-supervised formulation to learn node representations on graphs without any external labels.

\section{Methodology}
\begin{figure*}[t]
		\setlength{\belowcaptionskip}{-0.2cm}
		\centering
		{\includegraphics[width=1\textwidth]{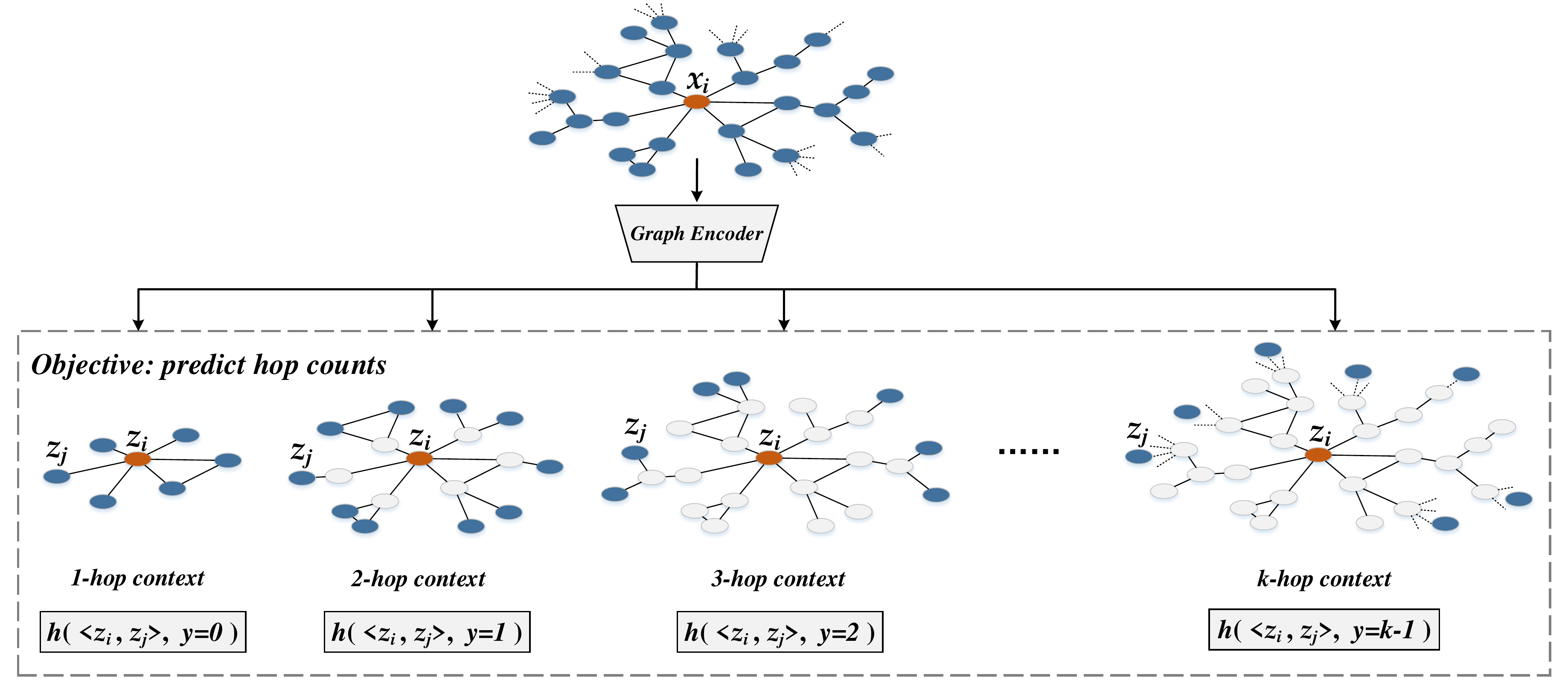}
			\caption{The proposed self-supervised framework S$^{2}$GRL for learning node representations over graph-structured data.
			}\label{fig:g2}}
\end{figure*}

\subsection{Problem Formulation}
Let $\mathcal{G}=(\mathcal{V}, \mathcal{E}, \mathcal{F})$ be a graph with $|\mathcal{V}|= n$ nodes and $|\mathcal{E}|= m $ edges where each node $v_{i} \in \mathcal{V}$ is affiliated with a set of $d$-dimensional attributes (features) $\mathcal{F} = \{f_{1}, f_{2}, \cdots, f_{d}\}$. $\Mat{X} = [\Vec{x}_{1}, \Vec{x}_{2}, \cdots, \Vec{x}_{n}]^{\top} \in \mathbb{R}^{n \times d}$ records the attribute information of $n$ nodes, where $\Vec{x}_{i}$ represents a feature vector of node $v_{i}$ for $i=1,2,\cdots, n$. Moreover, nodes are interconnected to form edges, represented by an adjacency matrix $\Mat{A} \in \{0,1\}^{n \times n}$. We aim to learn an encoder $f_{\omega}$ that projects each node to a $q$-dimensional space $\mathbb{R}^{q}$ under the guidance of natural supervision automatically obtained from the input graph itself instead of external annotated labels, such that the nodes will be represented in the global context as $\Mat{Z} = [\Vec{z}_{1}, \Vec{z}_{2}, \cdots, \Vec{z}_{n}]^{\top} \in \mathbb{R}^{n \times q}$. Formally, this free supervisory signals would function as pseudo-labels $\hat{Y}$ to train encoder $f_{\omega}$ by solving 
\begin{equation}\label{eq1}
    \min_{\omega,\theta}\mathcal{L}(\hat{Y},h_{\theta}(f_{\omega}(\Mat{X},\Mat{A}))),
\end{equation}
where $h_{\theta}$ is a classifier to predict the pseudo-labels. Note that labels $\hat{Y}$ are important in the learning procedure since they determine what should be captured and represented in latent vectors. In this sense, it is possible for us to construct specific pseudo-labels $\hat{Y}$ such that desired information can be encoded in node representations.


\subsection{Global Context of A Node}
Many studies~\cite{perozzi2014deepwalk,grover2016node2vec,qiu2018network} have found that the interaction between nodes is not limited to their direct connection, \emph{i.e.}, the observed first-order proximity, and the complementary high-order relations could capture more underlying information that would deepen our understanding of graphs. Thus, we assume that all nodes in $\mathcal{G}$ constitute the global context of node $v_{i}$ as any other node $v_{j} \in \mathcal{G}$ could interact with it through a path $p_{ij}$, which is much more comprehensive than a context specified by the limited window size in random walk based algorithms. Formally, such a global context of $v_{i}$ is denoted as $\mathcal{C}_{i} = \mathcal{V} -\{v_{i}\}$. 
To encode the global information, we plan to estimate the likelihood of predicting its context given one arbitrary node in $\mathcal{G}$, \emph{i.e.},
\begin{equation}\label{eq2}
    {\rm Pr}\,(\,\mathcal{C}_{i}\ |\ v_{i}).
\end{equation}
To learn representations, we introduce graph encoder $f_{\omega}$ into Eq.~\eqref{eq2} which presents a probability distribution of node co-occurrence, then yielding an optimization problem of maximizing the log-probability
\begin{equation}\label{eq3}
   \max_{\omega}\ \  \sum_{v_{i}\in \mathcal{V}}\log {\rm Pr}\,(\,\mathcal{C}_{i}\ |\ f_{\omega}(v_{i})).
\end{equation}

Then we factorize the objective function of optimization problem~\eqref{eq3} based on an independence assumption~\cite{grover2016node2vec} as following,
\begin{equation}\label{eq4}
    {\rm Pr}\,\left(\,\mathcal{C}_{i}\ |\ f_{\omega}\left(v_{i}\right)\right) = \prod_{v_{j} \in \mathcal{C}_{i}} {\rm Pr}\,\left(v_{j}\ |\ f_{\omega}\left(v_{i}\right)\right).
\end{equation}
For the conditional likelihood of each node pair $v_{i}$ and $v_{j}$, a typical solution is to define it as a softmax function
\begin{equation}\label{eq5}
    {\rm Pr}\,\left(v_{j}\ |\ f_{\omega}\left(v_{i}\right)\right) = \frac{{\rm exp}\left(f_{\omega}\left(v_{j}\right)^{\top}f_{\omega}\left(v_{i}\right)\right)}{\sum_{u \in \mathcal{V}}{\rm exp}\left(f_{\omega}\left(u\right)^{\top}f_{\omega}\left(v_{i}\right)\right)}, 
\end{equation}
and then adopt specific classifiers to learn such a posterior distribution, \emph{e.g.}, employing logistic regression to predict the context~\cite{mikolov2013distributed}. But such models would result in a large number of categories that equals to $|\mathcal{V}|$, consuming vast computational resource. Moreover, under our global context assumption, the classifier cannot be trained to work properly as all target classes will be positive. Hence, we propose  another strategy introduced as follows to predict ${\rm Pr}\,\left(v_{j}\ |\ f_{\omega}\left(v_{i}\right)\right)$, by utilizing hop count as supervision to guide the global context prediction in a fine-grained way.  

\subsection{A Natural Supervisory Signal: Hop Count}
An interesting discovery in the famous small-world experiment~\cite{newman2018networks} presents a heuristic: any pair of entities in the network owns the shortest path between them, which is usually the best path for message propagation. 
This makes it possible to divide the global context based on the hop count of the shortest path. Accordingly, we define a hop-based global context $\mathcal{C}_{i}$ for each node $v_{i}$ as follows.
\begin{definition}
A hop-based global context $\mathcal{C}_{i}$ includes a set of nodes that can reach $v_{i}$ through \textbf{the shortest path} with different hop counts, i.e., $\mathcal{C}_{i}$ is composed of multiple specific $k$-hop context $\mathcal{C}_{i}^{k}$ which only contains nodes within $k$ hops. 
\begin{equation*}
\begin{split}
    & \quad \mathcal{C}_{i} =\mathcal{C}_{i}^{1}\cup\mathcal{C}_{i}^{2}\cup\cdots\cup\mathcal{C}_{i}^{\delta_{i}},\\
    and\ \  &\mathcal{C}_{i}^{k}=\{v_{j}\ |\ d_{ij}=k\}, \ \ k=1,2,\cdots,\delta_{i},
\end{split}
\end{equation*}
where $\delta_{i}$ is the upper bound of the hop count from other nodes to $v_{i}$ in graph $\mathcal{G}$, and $d_{ij}$ is the length of path $p_{ij}$.
\end{definition}
For each target node $v_{i}$, node $v_{j} \in \mathcal{C}_{i}$ only belongs to one specific $k$-hop context $\mathcal{C}_{i}^{k}$, \emph{i.e.}, $\forall k_{1},k_{2} \in \{1,2,\cdots,\delta_{i}\}$, there exists $\mathcal{C}_{i}^{k_{1}}\cap\ \mathcal{C}_{i}^{k_{2}}=\emptyset$. It is not hard to find that such hop-based context can well reflect the extent of interaction between nodes $v_{i}$ and $v_{j}$. Specifically, if path $p_{ij}$ between $v_{i}$ and $v_{j}$ is relatively long, their communication has to pass through many relay points, and naturally, they interact with each other in a low extent. The closer the $k$-hop context to $v_{i}$ is, the stronger the relation between them. As a matter of course, hop counts can be utilized as supervisory signals to distinguish the degree of interaction between two nodes. 

In this way, for each target node $v_{i}$, with $\delta_{i}$ categories and pseudo-labels $Y_{i}=\{0,1,\cdots,\delta_{i}-1\}$ in accordance with $\mathcal{C}_{i}$, the learning objective, illustrated in Figure~\ref{fig:g2}, is to predict the hop count (relative contextual position) between $v_{j} \in \mathcal{C}_{i}$ and $v_{i}$ by solving the following optimization problem
\begin{equation}\label{eq6}
    \min_{\omega,\theta}\sum_{v_{j} \in \mathcal{C}_{i}}\mathcal{L}\left(Y_{i},h_{\theta}\left(\langle f_{\omega}\left(v_{i}\right)\, ,\,f_{\omega}\left(v_{j}\right)\rangle\right)\right),
\end{equation}
where $\langle\cdot\, ,\,\cdot\rangle$ is an operation used to measure the interaction between two vectors, and one of the available schemes is explained in~\textsection~\ref{sec:exp}. It can be seen that Eq.~\eqref{eq6} empowers the model to learn on unlabeled data in a supervised fashion. With the guidance of the pseudo-labels, the model not only encodes the global topology but also distinguishes the fine-grained interaction between nodes such that the learned representations are capable of finely characterizing the similarity and dissimilarity between nodes, which facilitates downstream tasks such as classification and clustering. 

Note that Eq.~\eqref{eq6} is difficult to solve in practice as the upper bound of hop count $\delta$ for different target nodes varies and precisely determining $\delta$ is not easy for a big graph. Therefore, modifying it to adapt to realistic situations is necessary. Inspired by the small-world phenomenon which demonstrates that two entities in the network are about six or fewer connections away from each other, we suppose that the number of hops between nodes is within a certain range, not uncontrollable (Note that the average shortest path length of the largest component in the tested benchmarks Cora, Citeseer, and Pubmed is 6.3, 9.3, and 6.3, respectively). So for $\delta_{i}$ classes attached to each node $v_{i}$, we divide them into $\alpha$ ``major'' categories by merging multiple classes, and update pseudo-labels to $\hat{Y} = \{0,1,\cdots,\alpha-1\}$ accordingly. Then, we obtain the objective function of the proposed S$^{2}$GRL:
\begin{equation}\label{eq7}
    \min_{\omega,\theta}\sum_{v_{i} \in \mathcal{V}}\sum_{v_{j} \in \mathcal{C}_{i}} \mathcal{L}\left(\hat{Y},h_{\theta}\left(\langle f_{\omega}\left(v_{i}\right)\, ,\,f_{\omega}\left(v_{j}\right)\rangle\right)\right).
\end{equation}
The design of ``major'' categories follows a principle of both clearly discriminating the dissimilarity and partly tolerating the similarity. For instance, the degree of interaction between a node and its 1-hop context is significantly different from its 2-hop context since you may not know your friends' friends at all. So treating them as two ``major'' classes is appropriate. In contrast, the distinction between higher-hop contexts is relatively vague, merging them into one ``major'' class seems more reasonable. $\alpha$ reflects the number of such predefined classes. A trivial solution is to treat the 1-hop context as one class and the rest as another, which is similar to the idea of reconstructing the adjacency matrix $\Mat{A}$. Detailed explanation about ``major'' categories explored in this work is in the following section. Now, the parameters of encoder $f_{\omega}$ and classifier $h_{\theta}$ can be jointly learned by optimizing Eq.~\eqref{eq7}, and the output of the optimized $f_{\omega}$ is our desired representation. 

\section{Experiments}

 
\subsection{Datasets}
We utilize a variety of standard real-world datasets collected from different domains to comprehensively evaluate the effectiveness of our S$^{2}$GRL on three common learning tasks. The detailed statistics are summarized in Table~\ref{tab:datasets}.

\begin{table}
	\tabcolsep= 3.5 pt
	\centering
	\caption{Dataset Statistics.}
	\begin{tabular}{ccccc}  
		\toprule
		\textbf{Dataset}		& \textbf{\# Nodes}  & \textbf{\# Edges}	& \textbf{\# Features} 	& \textbf{\# Classes}\\
		\midrule
		Cora		& 2,708 	& 5,429 	& 1,433 		& 7 		\\
		Citeseer	& 3,327 	& 4,732 	& 3,703 		& 6 		\\		
		Pubmed		& 19,717 	& 44,338 	& 500 			& 3 		\\
		PPI			& 56,944 	& 818,716 	& 50 			& 121 		\\
		Reddit		& 231,443 	& 11,606,919 & 602		 	& 41 		\\
		BlogCatalog	& 5,196 	& 171,743 	& 8,189 		& 6 		\\
		Flickr		& 7,575 	& 239,738 	& 12,047 		& 9 		\\
		\bottomrule
	\end{tabular}
	\label{tab:datasets}
\end{table}

\begin{itemize}
	\item \textbf{Cora, Citeseer, and Pubmed}~\cite{kipf2016semi}: three standard citation networks in which nodes are documents and edges indicate citation relations. In the experiments, they are employed for node classification (transductive) and clustering tasks.
	\item \textbf{PPI}~\cite{zitnik2017predicting}: a protein-protein interaction dataset that consists of networks corresponding to different human tissues.  It is used for node classification (inductive, multi-label) task.
	\item \textbf{Reddit}~\cite{hamilton2017inductive}: a social network constructed with Reddit posts in different topical communities. It is used for node classification (inductive) task.
	\item \textbf{BlogCatalog and Flickr}~\cite{li2015unsupervised}: two social networks in which users are treated as nodes and friend relations represent edges. Following~\cite{grover2016node2vec}, we randomly delete 20$\%$, 50$\%$, and 70$\%$ edges in these datasets, and use the damaged graph to conduct link prediction. 
\end{itemize}

\begin{table}
	\centering
	\tabcolsep= 5 pt
	\caption{Accuracy (\%) on transductive classification task.}
	\label{tab:cla_trans}
	\begin{tabular}{clccc}
		\toprule
		&\textbf{Algorithm} 	 & \textbf{Cora}		  & \textbf{Citeseer}		& \textbf{Pubmed}	\\
		\midrule
		\multirow{7}{*}{\rotatebox{90}{\small\textbf{label $\times$}}}& Raw features			& 56.6$\pm$0.4 		 & 57.8$\pm$0.2 	  	  & 69.1$\pm$0.2  \\
		&node2vec				 & 67.4$\pm$0.4 	  & 47.5$\pm$0.3 	   	   & 72.6$\pm$0.5  \\
		\cmidrule{2-5}
		&EP-B						 & 78.1$\pm$1.5 	 &71.0$\pm$1.4 	  & 79.6$\pm$2.1 	\\
		&DGI							 & 82.3$\pm$0.6 	 &71.8$\pm$0.7 	  & 76.8$\pm$0.6 	\\
		&graphite				  & 82.1$\pm$0.06 	  & 71.0$\pm$0.07 & 79.3$\pm$0.03 	\\
		&GMNN-unsup		   & 82.8 						& 71.5 					& 81.6 	\\
		&\textbf{S$^{2}$GRL} (ours)	& \textbf{83.7$\pm$0.2 }	& \textbf{72.1$\pm$0.5 }	& \textbf{82.4$\pm$0.2 }	\\
		\midrule
		\multirow{4}{*}{\rotatebox{90}{\small\textbf{label \checkmark}}}& GCN							& 81.5 					& 70.3 				   & 79.0 	\\
		&GAT							& 83.0$\pm$0.7 	& 72.5$\pm$0.7 	& 79.0$\pm$0.3 	\\
		&GWNN					 & 82.8 				 & 71.7 				& 79.1 	\\
		&GMNN-sup			 & 83.7 				 & 73.1 				& 81.8 	\\
		\bottomrule
	\end{tabular}
\end{table}

\subsection{Baseline Methods}
As our setup belongs to unsupervised learning, we mainly compare against two classes of the state-of-the-art unsupervised methods: random-walk based algorithms and GNNs. For the first category, we choose DeepWalk~\cite{perozzi2014deepwalk} and node2vec~\cite{grover2016node2vec}. For the latter, we select EP-B~\cite{duran2017learning}, DGI~\cite{velickovic2018deep}, graphite~\cite{grover2018graphite}, GMNN~\cite{pmlr-v97-qu19a}, and unsupervised GraphSAGE~\cite{hamilton2017inductive}. Particularly, since S$^{2}$GRL considers global topology, we also compare it with AGC~\cite{zhang2019attributed} which exploits adaptive graph convolution to capture high-order relations between nodes. To further demonstrate the potential of unsupervised learning, we give some additional results of supervised approaches, including GCN~\cite{kipf2016semi}, GAT~\cite{velivckovic2017graph}, FastGCN~\cite{chen2018fastgcn}, GWNN~\cite{xu2018graph}, and Adapt~\cite{huang2018adaptive}.

For fair comparison, the dimensionality of learned representations on all datasets is set to 512, unless noted otherwise. For node2vec, we set the number of random walks to 10, the walk length to 80, the window size to 10, and the parameters $p$ and $q$ both to 0.25. Parameters of DGI are the same as in~\cite{velickovic2018deep}. The results of other baselines are taken from their original papers.

\subsection{Experimental Setup}\label{sec:exp}

\begin{figure*}
	\centering
	\subfloat[DGI]{
		\includegraphics[width=0.195\linewidth]{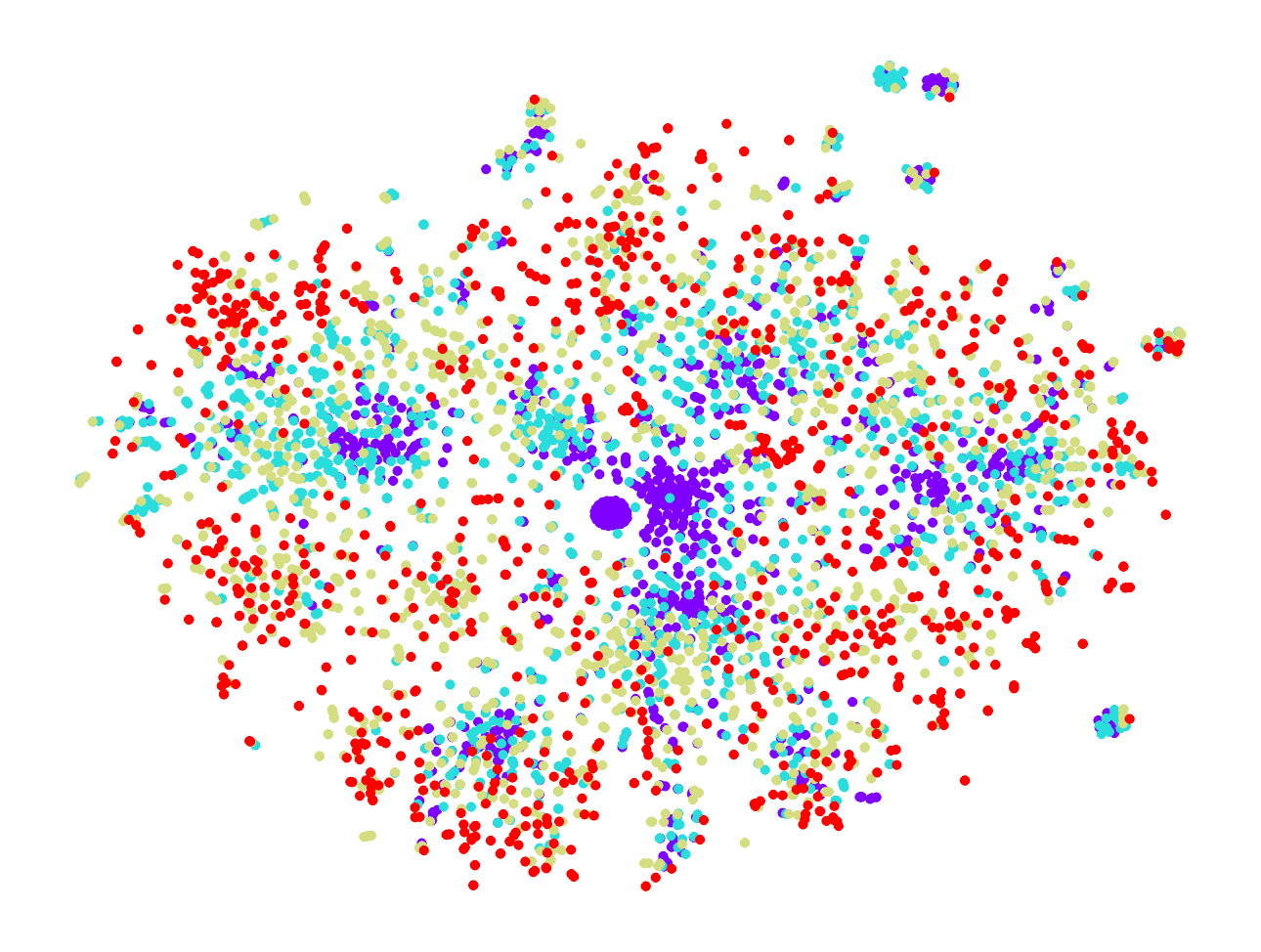}}
	\subfloat[S$^{2}$GRL (ours)]{
		\includegraphics[width=0.195\linewidth]{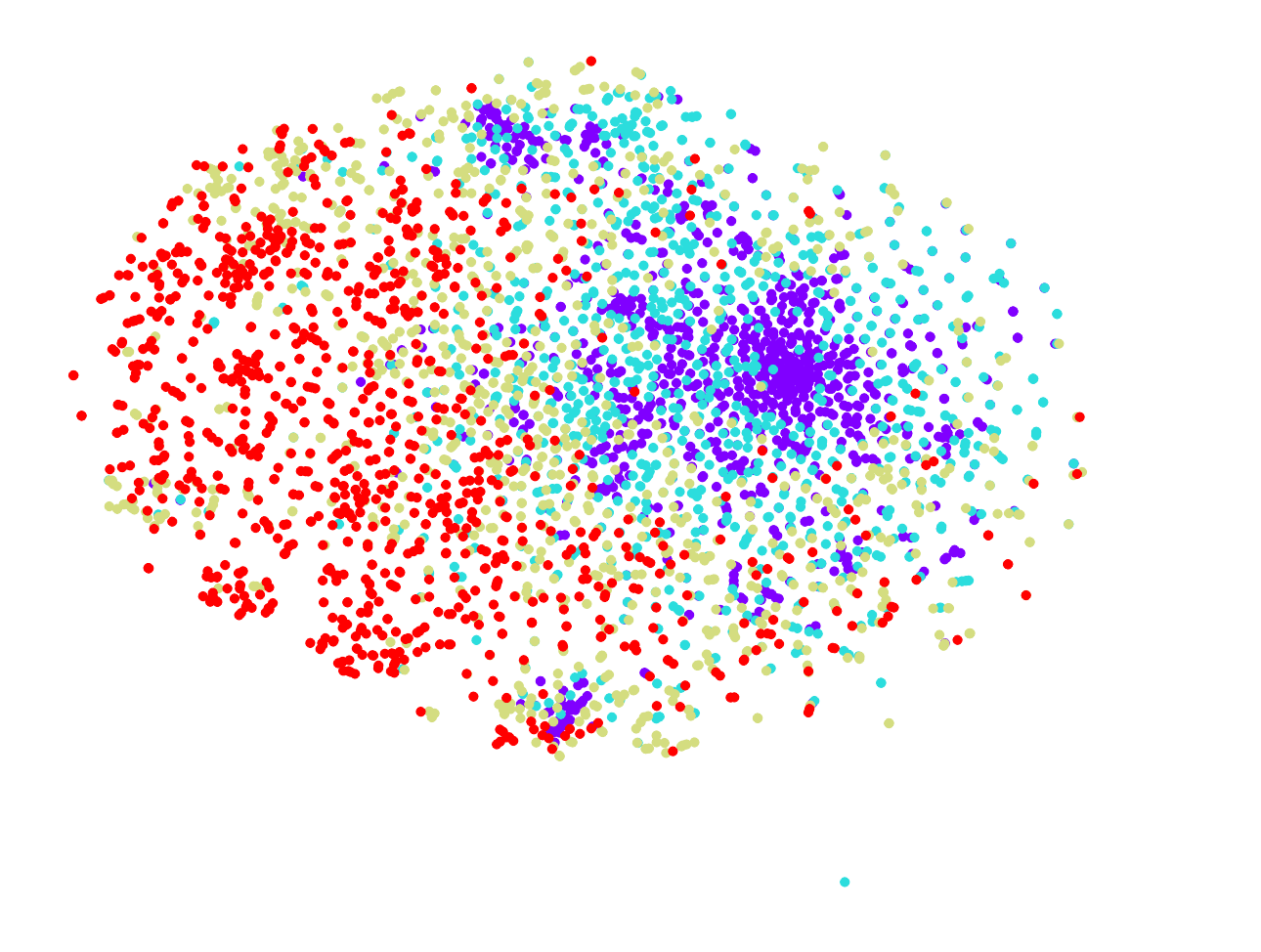}}
	\subfloat[node2vec]{
		\includegraphics[width=0.195\linewidth]{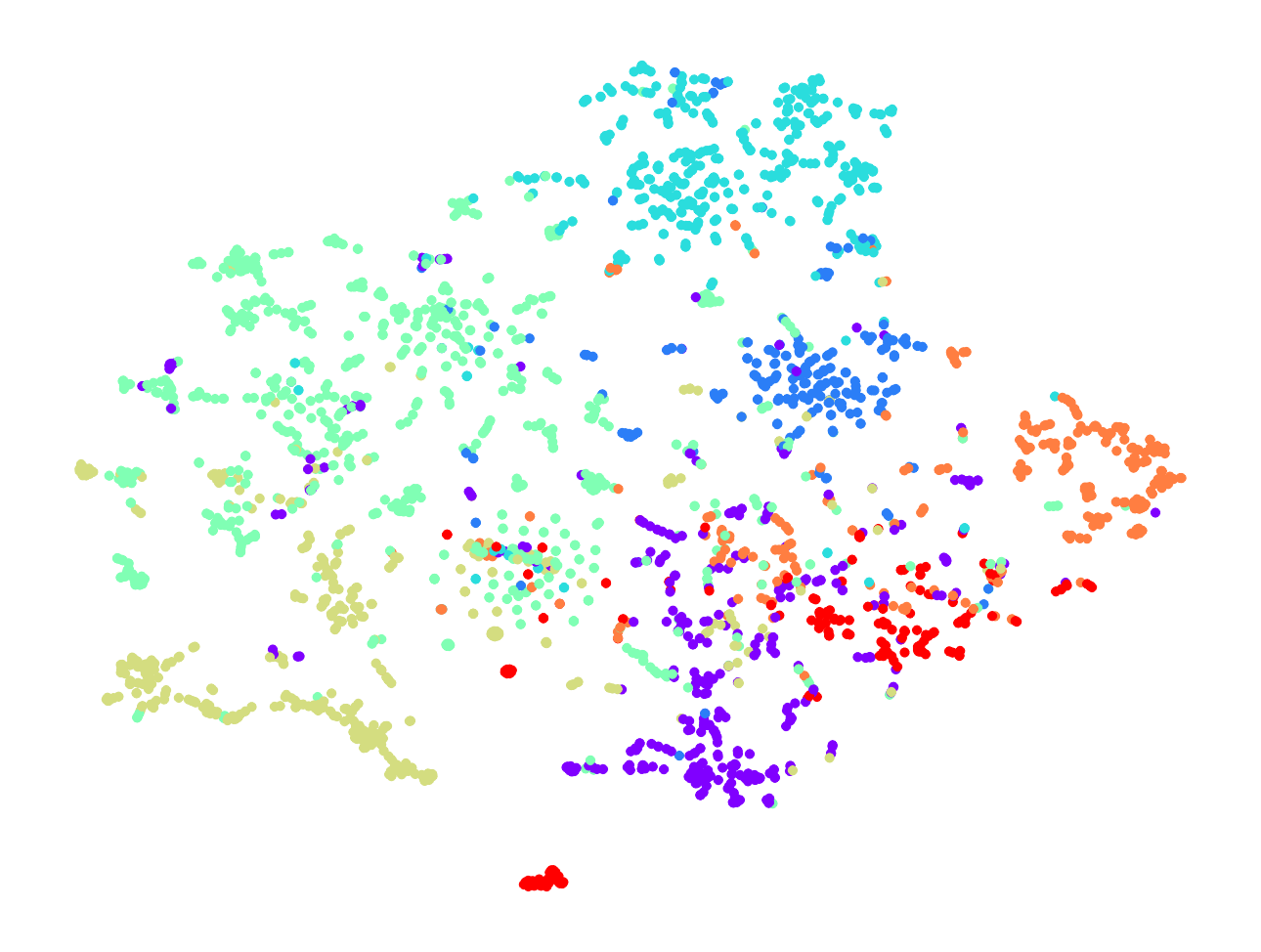}}
	\subfloat[DGI]{
		\includegraphics[width=0.195\linewidth]{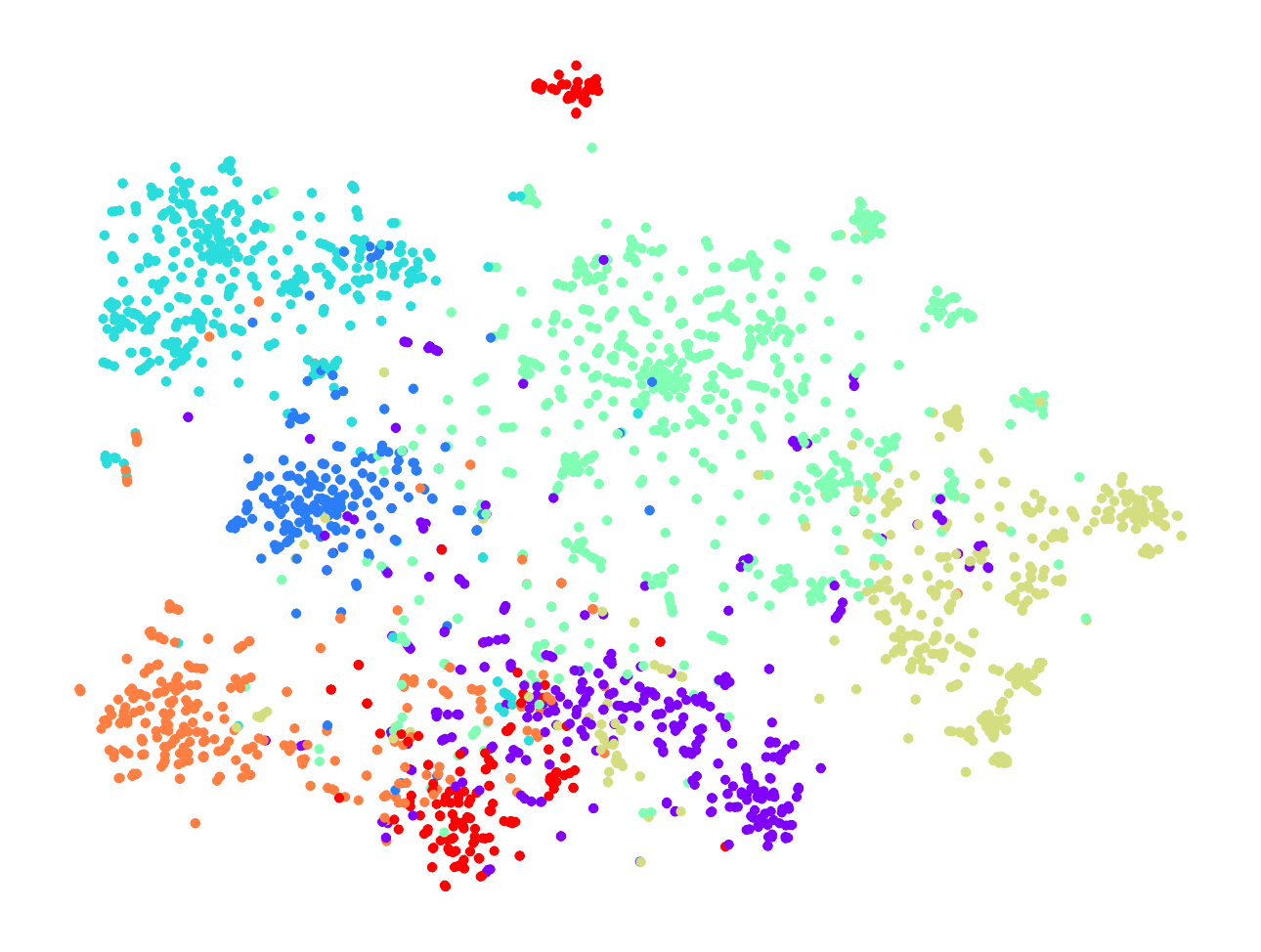}}
	\subfloat[S$^{2}$GRL (ours)]{
		\includegraphics[width=0.195\linewidth]{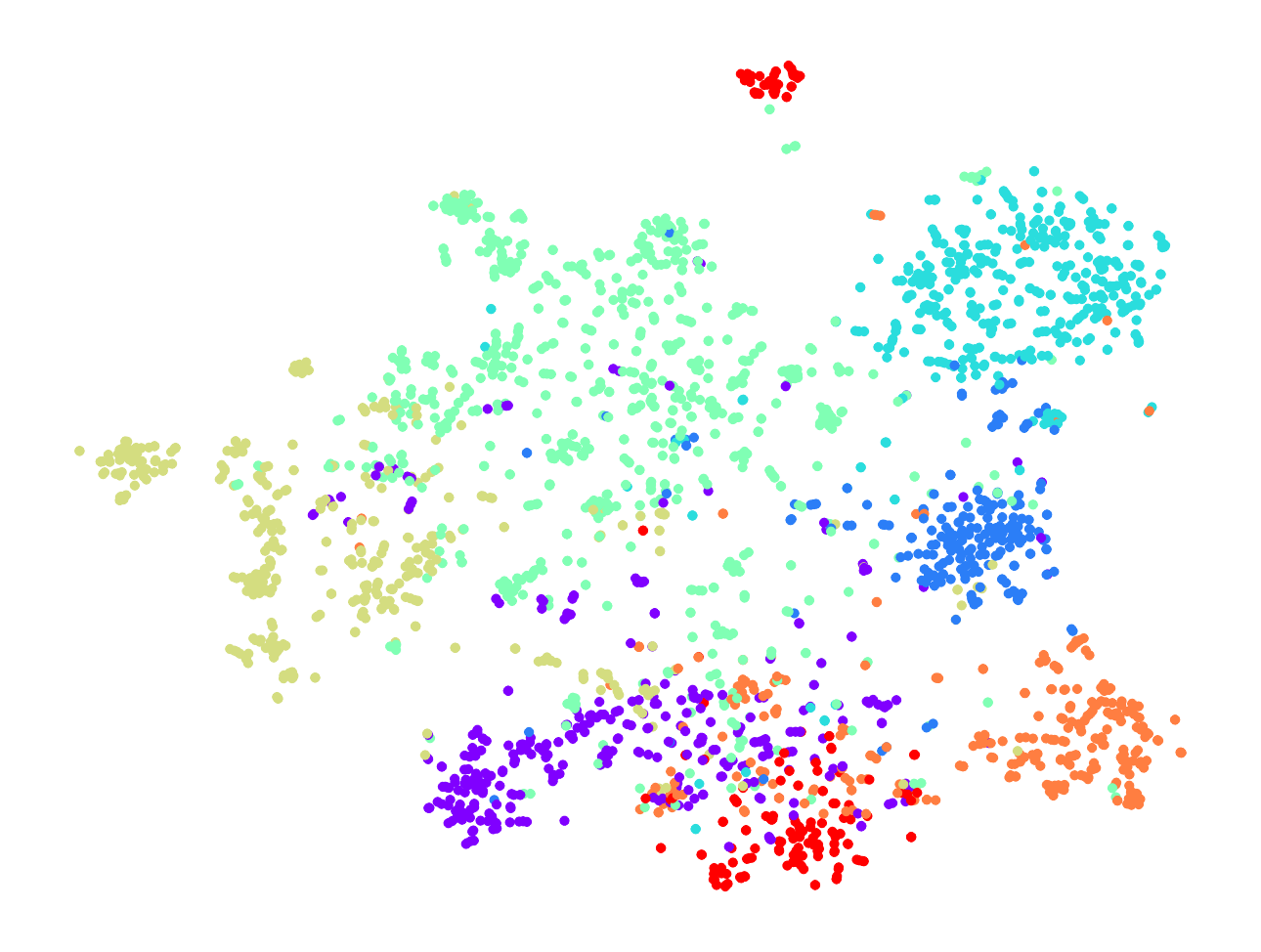}}
	\caption{\textbf{(a-b)} t-SNE plots of node pairs \emph{w.r.t.} topological distance on Cora. The color corresponds to the length of the shortest path between pairs of nodes. Vectors learned by S$^{2}$GRL present better structural properties.
	\textbf{(c-e)} Visualization of the learned representations on Cora. }
	\label{pic: vis}
\end{figure*}

\paragraph{Detailed architecture of S$^{2}$GRL.} For graph encoder $f_{\omega}$, we resort to the standard graph convolutional (GC) layer with ReLU activation function~\cite{kipf2016semi}. Specifically, in the inductive classification task (PPI and Reddit), we construct the encoder with two 512-neuron GC layers; in other tasks, our encoder is a 512-neuron GC layer. Considering that the $\langle\cdot\, ,\,\cdot\rangle$ used to measure the interaction between pairs of nodes should be symmetric to ensure $\langle \Vec{z}_{i}, \Vec{z}_{j}\rangle = \langle \Vec{z}_{j}, \Vec{z}_{i}\rangle$, we would calculate the element-wise distance between two vectors to achieve it, \emph{i.e.}, let $\langle \Vec{z}_{i}, \Vec{z}_{j}\rangle = abs(\Vec{z}_{i} - \Vec{z}_{j})$, where $abs(\cdot)$ means to take absolute value for each element. In addition, the following 4 ``major'' categories are used in our experiments: $\{\mathcal{C}_{i}^{1}, \mathcal{C}_{i}^{2}, \{\mathcal{C}_{i}^{3}\cup\mathcal{C}_{i}^{4}\}, \{\mathcal{C}_{i}^{5}\cup\cdots\cup\mathcal{C}_{i}^{\delta_{i}}\} \}$ (more discussion later).


\paragraph{Sampling strategy.} Note that Eq.~\eqref{eq7} involves the calculation between each pair of nodes in $\mathcal{G}$, which makes it computationally expensive and memory-consuming for large graphs. Besides, the number of node pairs included in each ``major'' class varies greatly, it would incur the class imbalance problem~\cite{japkowicz2002class}. To circumvent these issues, we perform a batch-sampling of node pairs based on a uniform distribution over the classes. In detail, We first randomly select a batch of fixed-size target nodes in $\mathcal{G}$, and then for each target node, sample node pairs from each `` major '' class at an adaptive ratio to ensure inter-class balance.


\begin{table}
	\centering
	\tabcolsep= 7 pt
	\caption{Micro-averaged F1 (\%) on inductive classification task.}
	\label{tab:cla_ind}
	\begin{tabular}{clcc}
		\toprule
		&\textbf{Algorithm}		& \textbf{PPI}		  & \textbf{Reddit}		\\
		\midrule
		\multirow{8}{*}{\rotatebox{90}{\small\textbf{label $\times$}}}&Raw features			   & 42.2					& 58.5					  \\
		&DeepWalk				   &   -					  & 32.4					 \\
		\cmidrule{2-4}
		&GraphSAGE-GCN		 & 46.5					  & 90.8					\\
		&GraphSAGE-mean		& 48.6					 & 89.7						\\
		&GraphSAGE-LSTM		& 48.2					 & 90.7						\\
		&GraphSAGE-pool		 & 50.2					  & 89.2					 \\
		&DGI						   & 63.8$\pm$0.20 & 94.0$\pm$0.10	 \\
		&\textbf{S$^{2}$GRL} (ours)	& \textbf{66.0$\pm$0.04}	& \textbf{95.0$\pm$0.02}	  \\
		\midrule
		\multirow{3}{*}{\rotatebox{90}{\small\textbf{label \checkmark}}}&GAT						  & 97.3$\pm$0.20		& -							\\
		&FastGCN				   & -							    & 93.7			   \\
		&Adapt						 & -							  & 96.3$\pm$0.32  \\
		 \bottomrule
	\end{tabular}
\end{table}

\paragraph{Implementation details.} As a preprocessing step, we employ NetworkX to build the hop-based global context for each node in parallel.
During training, we adopt Glorot initialization~\cite{glorot2010understanding} and Adam optimizer~\cite{kingma2014adam} with an initial learning rate in $\{0.001, 0.003, \dots, 0.009\}$. The number of epochs is tuned in $\{100, 200, 300\}$, while setting a fixed number of epochs on the inductive classification task (20 on Reddit, 50 on PPI). Besides, we use the subsampling skill introduced in~\cite{hamilton2017inductive} to make Reddit and PPI fit into GPU memory. In specific, a minibatch of 256 nodes is first selected, and then for each selected node, we uniformly sample 8 neighbors from its first and second-order neighborhoods, respectively.

\paragraph{Evaluation metrics.} Following the experimental setup described in~\cite{velickovic2018deep}, we feed the learned representations into a simple logistic regression classifier to evaluate the node-level classification performance. Mean accuracy
after 50 runs is used to assess the transductive task, and the micro-averaged F1 score averaged after 50 runs is used for the inductive one. For the clustering task, we apply the K-means algorithm to group the learned embeddings and report the NMI score. For the link prediction task, we use Area Under the ROC Curve (AUC) as the criteria. Similarly, we report the averaged result after 10 runs.

\subsection{Results}

\paragraph{Node classification.} The results of transductive and inductive task are reported in Tables~\ref{tab:cla_trans} and~\ref{tab:cla_ind}, where numbers in bold indicate the best results among unsupervised methods. As can be observed, S$^{2}$GRL outperforms all other unsupervised algorithms, especially on Pubmed and PPI, which affirms the effectiveness of hop count as free supervisory signals. This confirms the benefit of our proposed self-learning task, \emph{i.e.}, global context prediction. Good performance on reasoning about the relative contextual position can only be achieved if the learned representations could encode global topological information and finely discriminate the similarity and differentiation between nodes (t-SNE~\cite{maaten2008visualizing} plots \emph{w.r.t.} topological distance are given in Figure~\ref{pic: vis} (a-b)), which indirectly contributes to classification. Besides, S$^{2}$GRL exhibits comparable results to some supervised models like GCN and GWNN, even achieves the best result on Pubmed. We believe that self-supervised learning has more potential in learning high-quality representations than supervised manners as the supervision built from the data itself could capture the inherent characteristics of data better than manual labels. Moreover, S$^{2}$GRL exhibits a comparable training speed with GNN-based baselines.



\paragraph{Clustering.} Table~\ref{tab:clu} summarizes the results. Although DGI achieves the best performance on Cora and Citeseer, our simple framework S$^{2}$GRL also exhibits competitive performance (an illustration is shown in Figure~\ref{pic: vis} (c-e)) and obtains the highest NMI on Pubmed. Note that S$^{2}$GRL outperforms AGC, a clustering-oriented model adaptively capturing high-order relations among nodes, which demonstrates that high-order relations are somewhat limited in capturing the underlying structures of the graph while our consideration of global topology and fine-grained similarity can be beneficial to learn cluster structures. 


\begin{table}
	\centering
	\caption{Clustering quality in terms of NMI.}
	\label{tab:clu}
	\begin{tabular}{lccc}
		\toprule
		\textbf{Algorithm} 	 & \textbf{Cora}		  & \textbf{Citeseer}		& \textbf{Pubmed}	\\
		\midrule
		Raw features			& 0.135						& 0.237						   & 0.314					  \\
		node2vec				 & 0.449					 & 0.232						& 0.288					   \\
		\midrule
		DGI							 & \textbf{0.557}					 & \textbf{0.438}						& 0.292					  \\
		AGC							& 0.537   	& 0.411    	   & 0.316					 \\
		\midrule
		\textbf{S$^{2}$GRL} (ours) & 0.540				  & 0.432						 & \textbf{0.332}		\\
		\bottomrule
	\end{tabular}
\end{table}

\begin{table}
	\centering
	\tabcolsep =4.0pt
	\caption{AUC scores (\%) for link prediction.}
	\label{tab:lp}
	\begin{tabular}{lcccccc}
		\toprule
		\multirow{2}{*}{\textbf{Algorithm}}	 & \multicolumn{3}{c}{\textbf{BlogCatalog}} 		  & \multicolumn{3}{c}{\textbf{Flickr}}	\\
		& \textbf{20\%}		& \textbf{50\%}		& \textbf{70\%}		& \textbf{20\%}	& \textbf{50\%}	& \textbf{70\%}	\\
		\midrule
		node2vec	& 79.9		& 76.5		& 72.4		 & 73.9		& 70.0 		& 63.1		\\
		DGI				& 77.7		& 76.0		& 75.4		 & 90.6		& 88.8		& 69.2		\\
		\midrule
		\textbf{S$^{2}$GRL} (ours) & \textbf{80.4}	 & \textbf{78.7}		  & \textbf{78.2}	 & \textbf{91.4}		 & \textbf{90.9}	& \textbf{89.8}	\\
		\bottomrule
	\end{tabular}
\end{table}

\paragraph{Link prediction.} As can be seen from Table~\ref{tab:lp}, S$^{2}$GRL consistently outperforms DGI and node2vec under different edge removal rates, showing that the representation learned by global context prediction could delicately characterize the similarity and differentiation between nodes from a global topological viewpoint to predict missing links. By contrast, the necessity of a task-oriented negative sample generating function weakens the performance of DGI in this task. This shows that S$^{2}$GRL has a better generalization ability.


\subsection{Further Discussion on Label Categories}
In Table~\ref{tab:lc} we investigate how the quality of the self-supervised learned embeddings depends on the construction of ``major'' classes. It can be found that clearly distinguishing 1-hop, 2-hop, and 3-hop contexts into 3 distinct ``major'' classes benefits to improving the quality of representations, while further differentiating 4-hop and higher-hop contexts would degrade the performance. We believe the reason is that only making 1-hop context discernible offers too few categories for recognition, \emph{i.e.}, providing less supervisory information, while too many ``major'' categories are not distinguishable enough as the distinction between higher-hop contexts is vague. Hence, in the experiments, we adopt a scheme that combines 3-hop and 4-hop contexts into a single class, which indeed presents better empirical performance.

\begin{table}
	\centering
	\caption{Accuracy (\%) \emph{w.r.t.} the formation of label classes on Cora.}
	\label{tab:lc}
	\begin{tabular}{clc}
		\toprule
		\textbf{\# Classes} & \textbf{Merge Policy} & \textbf{Accuracy}\\
		\midrule
		 2& $\mathcal{C}_{i}^{1}, \mathcal{C}_{i}^{2}\cup\cdots\cup\mathcal{C}_{i}^{\delta_{i}}$ & 82.4\\
		 3&$\mathcal{C}_{i}^{1}, \mathcal{C}_{i}^{2}, \mathcal{C}_{i}^{3}\cup\cdots\cup\mathcal{C}_{i}^{\delta_{i}}$&83.0\\
		 4&$\mathcal{C}_{i}^{1}, \mathcal{C}_{i}^{2}, \mathcal{C}_{i}^{3}, \mathcal{C}_{i}^{4}\cup\cdots\cup\mathcal{C}_{i}^{\delta_{i}}$&83.2\\
		 5&$\mathcal{C}_{i}^{1}, \mathcal{C}_{i}^{2}, \mathcal{C}_{i}^{3}, \mathcal{C}_{i}^{4}, \mathcal{C}_{i}^{5}\cup\cdots\cup\mathcal{C}_{i}^{\delta_{i}}$&82.7\\
		 6&$\mathcal{C}_{i}^{1}, \mathcal{C}_{i}^{2}, \mathcal{C}_{i}^{3}, \mathcal{C}_{i}^{4}, \mathcal{C}_{i}^{5}, \mathcal{C}_{i}^{6}\cup\cdots\cup\mathcal{C}_{i}^{\delta_{i}}$&82.7\\
		\bottomrule
	\end{tabular}
\end{table}

\section{Conclusion}

In this work, we have presented a novel self-supervised framework S$^{2}$GRL for learning node representations, which to our knowledge is the first attempt on exploring free supervisory signals in graph-structured data for representation learning. Extensive experiments demonstrate the effectiveness of our framework. We hope our work will inspire more research in self-supervised graph representation learning.


\newpage
\bibliographystyle{named}
\bibliography{reference}

\end{document}